\documentclass{article}
\usepackage{ijcai26}

\usepackage{times}
\usepackage{soul}
\usepackage{url}
\usepackage[hidelinks]{hyperref}
\usepackage[utf8]{inputenc}
\usepackage[small]{caption}
\usepackage{graphicx}
\usepackage{amsmath}
\usepackage{amssymb}
\usepackage{makecell}
\usepackage{multirow}
\usepackage{subcaption}
\usepackage{amsthm}
\usepackage{booktabs}
\usepackage{algorithm}
\usepackage{hyperref}
\usepackage{comment}
\usepackage{algorithmic}
\usepackage[switch]{lineno}
\usepackage{eso-pic}

\title{Suicide Risk Assessment from AI-powered Video Surveillance: \\An Interpretable Framework for Prevention in Metro Stations}

\author{
Safwen Naimi$^1$
\and
Wassim Bouachir$^1$\and
Guillaume-Alexandre Bilodeau$^{2}$\And
Brian Mishara$^3$\\
\affiliations
$^1$Université TÉLUQ, QC, Canada\\
$^2$Polytechnique Montréal, QC, Canada\\
$^3$Université du Québec à Montréal, QC, Canada\\
\emails
\{safwen.naimi, wassim.bouachir\}@teluq.ca,
guillaume-alexandre.bilodeau@polymtl.ca,
mishara.brian@uqam.ca
}

\begin{document}

\AddToShipoutPictureFG*{%
  \AtPageUpperLeft{%
    \hspace{0.5\paperwidth}%
    \raisebox{-1.5cm}{%
      \makebox[0pt][c]{\normalsize \underline{Accepted for publication in the International Joint Conference of Artificial Intelligence (IJCAI)}}%
    }%
  }%
}

\maketitle

\begin{abstract}

Understanding and monitoring human behavior in metro stations play an important role in supporting suicide prevention efforts, where early identification of high-risk situations can enable timely intervention. This requires assessing suicide risk from a surveillance video by jointly reasoning about the behavior of each passenger, his/her spatial context, and temporal dynamics. However, this assessment using videos captured by surveillance cameras is challenging, as it demands accurate perception of human motion, understanding of platform geometry, and aggregation of heterogeneous behavioral cues over time. In this work, we formalize the task of Suicide Risk Assessment (SRA) in metro stations and introduce the first interpretable framework that addresses this challenge. Unlike approaches that focus on isolated subtasks or attempt to infer intent directly, our formulation assesses suicide risk from accumulated evidence by incorporating person tracking, activity recognition, semantic segmentation of the platform, and trajectory-driven risk heatmap modeling. By formalizing SRA as a distinct task and benchmarking a complete operational pipeline achieving 83.2\% ROC-AUC on real surveillance data, this work highlights the complexity of suicide risk assessment and opens new directions for research on interpretable AI systems for social good.

\end{abstract}

\section{Introduction}

Suicide remains a major public health concern worldwide \cite{Forte2021TheRO,Rassy2021InformationAC}. The deployment of surveillance cameras at known suicide hotspots and high-risk public spaces, such as metro stations, has created opportunities for monitoring and early intervention \cite{Anjali2024SpatialAO,Bouachir2018IntelligentVS}. Existing vision-based systems, however, largely depend either on continuous human observation or on simple visual cues such as detecting the crossing of safety barriers to trigger emergency responses \cite{Laliberte2021SuicideSA}. These strategies often detect incidents at a late stage, limiting their preventive impact, while remaining time-consuming and prone to human fatigue. Supporting human operators with automated systems that can highlight potentially high-risk situations in real time while remaining interpretable and ethically grounded has therefore become an important challenge at the intersection of artificial intelligence and social good \cite{Balamurugan2025HumansIT,Gursel2024TheRO}.

Video surveillance infrastructures are already widely deployed in metro stations, offering a rich source of visual information for understanding passenger behavior and movement patterns. Several prior studies have investigated the analysis of videos to better understand behaviors preceding suicide attempts in metro and railway environments \cite{Mishara2016CanCI,ParsapoorMahParsa2023SuicideRD,Rdbo2012PatternsOS,Silla2012MainCO,Mackenzie2018BehavioursPS}. These works, often grounded in interdisciplinary collaborations between psychology, sociology, and transportation authorities, have identified recurring behavioral and spatial patterns, such as repeated pacing near the far-end zone of the platform, prolonged standing or walking on the yellow platform demarcation line, and looking toward the tunnel. Comparative investigations between attempters and non-attempters have further shown that risk-related behaviors are not defined by a single distinctive action, but by patterns of accumulation and repetition over time \cite{Mishara2016CanCI}. 
We believe that developing reliable systems for suicide risk assessment in metro stations requires an interdisciplinary approach, and that a key step forward is the integration of knowledge from researchers in suicide prevention with artificial intelligence advancement for effective real-time video analysis. 

In this work, we formalize Suicide Risk Assessment (SRA) as a distinct computational task. Rather than predicting intent or labeling individuals, SRA aims to assess the level of suicide risk associated with observable behavior and spatial context over time for each passenger. This formulation aligns with interdisciplinary findings that view suicide risk as an emergent phenomenon, shaped by the interaction between individuals and their environment \cite{Hightower2024ProposingAI,Garipy2024DynamicSM}. We propose an interpretable framework that integrates multiple complementary modules capable of producing continuous, individual-level suicide risk assessment grounded in spatial exposure, temporal persistence, and behavioral patterns, while avoiding reliance on appearance cues or personal attributes.
\vspace{5pt}
\newline
\textbf{Interdisciplinary Collaboration.} This work is the result of an interdisciplinary collaboration between computer scientists (at universities), researchers in the field of suicidology, and The Société de transport de Montréal (STM).
\vspace{5pt}
\newline
\textbf{Contributions.} We summarize our contributions as follows: \textbf{(i)} We introduce \textbf{SRA-Framework}, the first interpretable SRA pipeline for suicide risk assessment in metro stations, \textbf{(ii)} We design a set of spatial, temporal, and behavioral risk indicators aligned with findings in suicide prevention and show how their aggregation over time enables effective and interpretable suicide risk assessment, and \textbf{(iii)} We evaluate the proposed framework on ethically approved, real-world surveillance video data from metro stations, demonstrating meaningful discrimination between at-risk and control scenarios and providing qualitative and quantitative interpretability analyses.

\section{Related Work}

\subsection{Behavioral Studies for Suicide Prevention in Metro Stations}
\label{sec2.1}
The study of behaviors preceding a suicide attempt in public transportation stations, particularly through surveillance cameras, is a crucial research area for proactive suicide prevention. Several studies identified the behavioral signs that precede such acts in order to enable rapid intervention \cite{Mishara2016CanCI,Rdbo2012PatternsOS,Silla2012MainCO,Mackenzie2018BehavioursPS}. These studies report that attempters spend significantly more time on the yellow line and the far-end zone of the platform, compared to control passengers. In addition to spatial proximity, repeated movement patterns, such as back-and-forth pacing between the wall of the platform and the yellow line, as well as repetitive looking at the tunnel, have been shown to occur more frequently among attempters. Temporal factors play an important role because at-risk situations are characterized by the accumulation and repetition of risk-related behaviors over time. According to Mishara \textit{et al.} \shortcite{Mishara2016CanCI}, 83\% of attempters exhibited easily observable behaviours potentially indicative of an impending attempt, and 61\% had two or more of these behaviours, while 75\% had at least one of these behaviours. According to the same study, two behaviours: leaving an object on the platform and going back and forth between the wall and the yellow line, could identify 24\% of attempters with no false positives. The other target behaviours could also be present in non-attempters, but with fewer occurrences. However, having two or more of these behaviours indicated a high likelihood of being at risk of attempting suicide.

\subsection{Deep learning-based abnormal behavior recognition in Metro Stations}

Deep learning approaches have been widely explored for abnormal behavior recognition from video surveillance, aiming to automatically identify unusual or potentially dangerous events in metro stations. These methods typically rely on spatio-temporal representations learned from raw video frames, optical flow, trajectories, or pose information, and are often formulated as anomaly detection or action classification problems. Naimi \textit{et al.} \shortcite{Naimi2025SSTARSS} introduced SSTAR, a skeleton-based action recognition model designed to recognize pre-suicidal passenger activities. While SSTAR demonstrates strong performance in identifying semantically meaningful actions under challenging surveillance conditions, it does not account for spatial context and risk assessment. 
Zuo \textit{et al.}~\shortcite{Zuo2025TheRO} formulates passenger abnormal behavior recognition as a video anomaly detection task, using a Swin Transformer-based generative model to reconstruct surveillance frames and detect abnormalities via reconstruction error, with additional motion refinement through optical flow. This approach achieves good performance but focuses on frame-level anomaly detection rather than interpretable, context-aware risk assessment. Chang \textit{et al.}~\shortcite{Chang2025UsingDL} proposes a smart railway safety monitoring system that integrates deep learning–based segmentation and object detection to identify intrusions on railway tracks. Platform boundaries are automatically extracted using Mask R-CNN \cite{He2017MaskR}, while YOLOv3 \cite{Redmon2018YOLOv3AI} is employed to detect human and object intrusions under diverse environmental conditions. This approach focuses on intrusion detection and alerting rather than behavioral analysis or suicide risk assessment. 
Different from previous works, which largely focus on anomaly or intrusion detection, our SRA-Framework formulates suicide prevention as a long-term, context-aware risk assessment problem. We integrate multiple vision components: tracking, platform segmentation, activity recognition, and spatial homography to extract interpretable behavioral and spatial indicators that are accumulated over time. This enables continuous suicide risk monitoring under real surveillance conditions rather than instantaneous event detection.

\section{Suicide Risk Assessment Task}

\begin{figure*}[t]
        \centering	\includegraphics[width=0.98\textwidth]{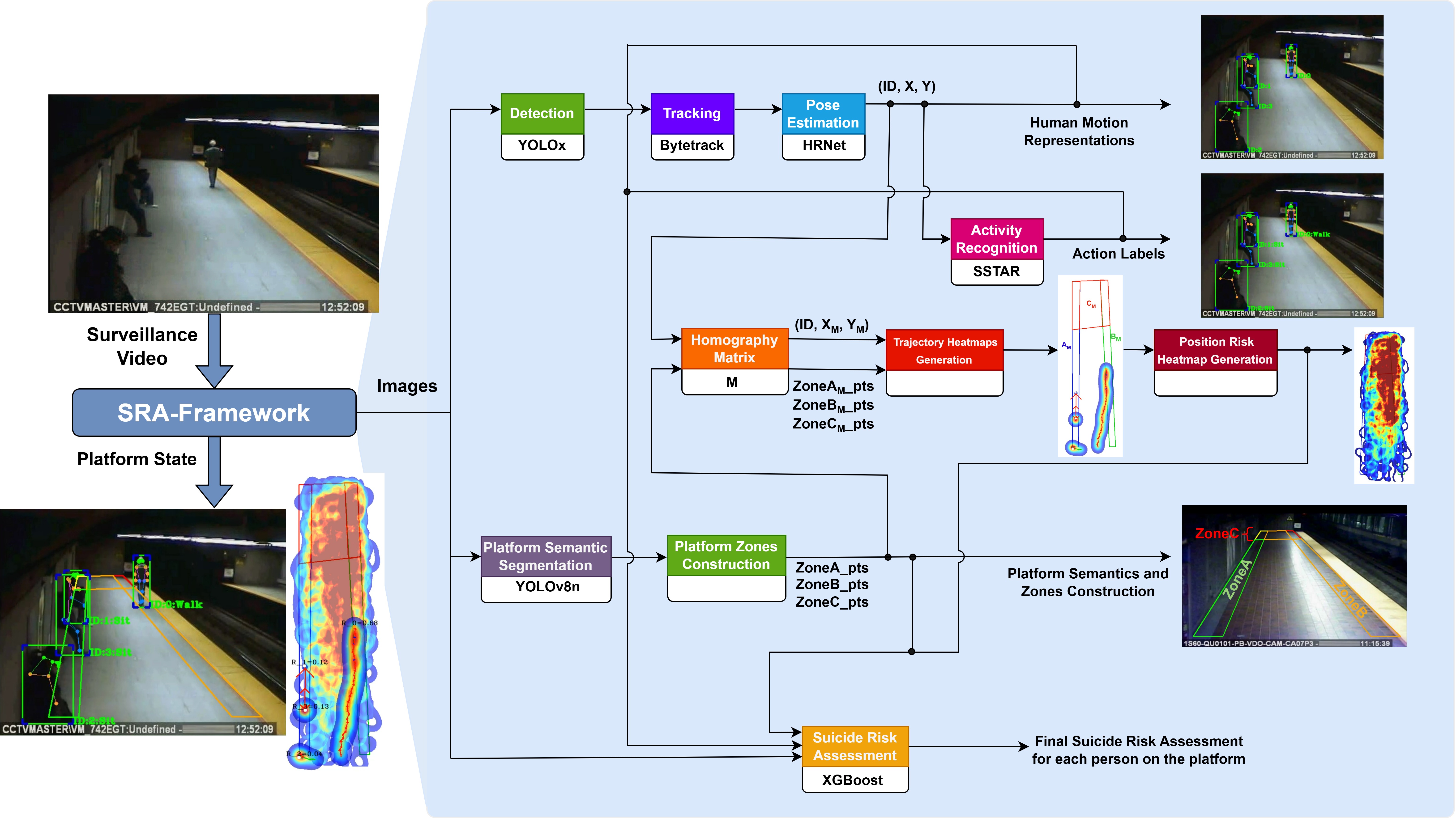}
	\caption{\textbf{Architecture overview of our proposed SRA-Framework.} It takes a video as input and outputs the complete platform state and a suicide risk assessment for each person.
}
	\label{FIG:1}
\end{figure*}

We address the problem of suicide risk assessment in metro stations using continuous surveillance video streams. The objective is to assist preventive monitoring by assessing, for each observed passenger, a time-varying risk score that reflects their level of suicide risk based on observable behavior and spatial context. 

Let $V=\left\{F_t\right\}_{t=1}^T$ denote a sequence of video frames acquired from a camera. At each time step $t$, the system detects and tracks a set of individuals $\mathcal{P}_t=\left\{p_i\right\}_{i=1}^{M_t}$. For each individual 
$p_i \in \mathcal{P}_t$, the task is to output a scalar suicide risk score $R_{p_i}(t) \in[0,1]$, where higher values indicate elevated suicide risk. Formally, suicide risk is modeled as a function of accumulated behavioral and spatial indicators extracted over a temporal window $W_t=[t-\tau+1, t]$ such as:
\begin{equation}
\label{eq:1}
 \quad R_{p_i}(W_t)=\sigma\left(f\left(\mathbf{x}_{p_i}(W_t)\right)\right),
\end{equation}

\noindent where $\mathbf{x}_{p_i}(W_t) \in \mathbb{R}^d$ denotes a feature vector summarizing $d$ observable behaviors and interactions between the individual and the platform environment over time, $f(\cdot)$ represents a supervised learning model, and 
$\sigma(\cdot)$ maps the output to the interval $[0,1]$. Note that $R_{p_i}(W_t)$ does not represent suicidal intent, a psychological state, or a clinical diagnosis. Instead, 
it quantifies suicide risk, defined as the degree to which an individual’s observable behavior and spatial interaction with different areas of the platform may warrant preventive attention. It is influenced not only by what actions are performed, but also by where they occur and how long they persist. This formulation is consistent with operational safety practices, where decisions are informed by the accumulation of behavioral and contextual cues rather than isolated events.

\section{Proposed SRA-Framework}

Given a continuous surveillance video stream, our framework outputs a platform state representing a time-varying suicide risk score $R_{p_i}(W_t) \in[0,1]$ for each tracked individual $p_i$. The overall architecture of our SRA-Framework is depicted in Fig. \ref{FIG:1}. 
First, human motion representations of each person on the platform are extracted using sequential detection, tracking, and pose estimation modules. Second, a skeleton-based activity recognition module uses pose dynamics to predict action labels of each person over short temporal windows. Third, a platform segmentation module captures the spatial structure of the station by segmenting the platform into relevant zones, which serve as a basis for identifying back-and-forth and repetitive movement behaviors. Fourth, a position risk heatmap generation module aggregates the platform-referenced heatmaps of multiple at-risk individuals over time, producing a position risk heatmap that reflects the most frequently traversed or occupied areas by persons at risk within the platform. Finally, a risk inference module aggregates per-person indicators into a feature vector $\mathbf{x}_{p_i}(t)$ and computes the suicide risk score $R_{p_i}(W_t)$. To simplify notation, in the following, we will denote an individual $p_i$ as $p$.

\subsection{Human Detection, Tracking, and Pose Estimation}
\label{sec4.1}
We employ a pre-trained YOLOx \cite{ultra} model as our human detector, without fine-tuning it on our dataset, since it already provides decent performance on human videos. We filter the model output to retain only the "person" class detections. 

To leverage existing strong multi-object trackers, our SRA-Framework performs tracking in the image space based on bounding boxes. We employ ByteTrack \cite{Zhang2021ByteTrackMT} as our multi-object tracker, for its good performance and its ability to leverage both spatio-temporal and appearance cues. For each tracked individual, a pretrained HRNet \cite{Wang2019DeepHR} model is applied to extract a set of 2D body keypoints within the corresponding bounding box. Pose estimates are accumulated over temporal windows to form pose sequences that capture fine-grained motion dynamics while remaining invariant to appearance and scale. This representation enables the system to reason about human movement patterns beyond coarse trajectory information.

As illustrated in Fig. \ref{FIG:1}, tracked positions are subsequently projected into a platform-referenced coordinate system using a homography matrix $M$, which enables the construction of trajectory-based heatmaps that encode where and how long each individual moves or remains within the platform.

\begin{figure*}[t]
        \centering	\includegraphics[width=1.0\textwidth]{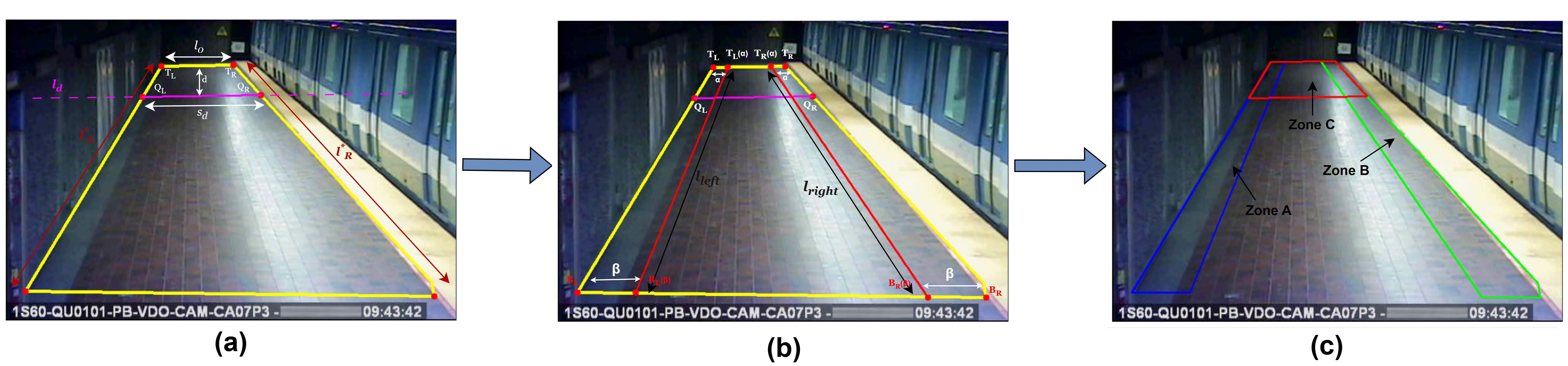}
	\caption{\textbf{Overview of the Platform Semantics Modeling Process:} (a–b) Boundary and Offset Lines Estimation, and (c) Final Zone Partitioning.}
	\label{FIG:2}
\end{figure*}

\subsection{Human Activity Recognition}
\label{sec4.2}

Given the tracked pose sequences extracted in section \ref{sec4.1}, this module infers human activity patterns over temporal windows $W_t$, providing high-level action cues for subsequent risk assessment. In our SRA-Framework, we employ SSTAR pretrained on the ARMM dataset \cite{Naimi2025SSTARSS}, a skeleton-based spatio-temporal action recognition model specifically designed for surveillance scenarios. In our case, SSTAR is used to detect 3 passenger action labels: ‘LookTunnel’, ‘Walk’, and ‘Stand’. 
The output of this module consists of per-person activity labels associated with each temporal window. Importantly, the contribution of recognized activities in our system depends on their spatial context, duration, and repetition. 

\subsection{Platform Semantics and Zones Construction}
\label{sec4.3}

To incorporate environmental context, we model platform semantics by partitioning the scene into spatial zones that support interpretable risk indicators, such as back-and-forth movements between the wall and the yellow line and explicit far-end zone entry events. Platform segmentation is obtained using YOLOv8n \cite{ultra} in its semantic segmentation configuration, chosen for its favorable accuracy–efficiency trade-off. The model is trained offline on manually annotated images with pixel-level labels of the platform surface.

For zones construction, we estimate the two lateral platform boundaries corresponding to the wall-facing side ($\ell_L^*$) and the track-facing side ($\ell_R^*$) of the platform using the segmentation result. As shown in Fig. \ref{FIG:2}a, let $\mathbf{T}_L$ and $\mathbf{T}_R$ represent the top left point and the top right point of the platform contour respectively. Given the top platform edge segment $\ell_o=\overline{\mathbf{T}_L \mathbf{T}_R}$, we generate an inward parallel line segment $\ell_d$ at offset distance $d$ using the unit normal vector. To ensure geometric validity within the platform extent, $\ell_d$ is clipped by intersecting it with the estimated left and right platform boundary line segments $\ell_L^*$ and $\ell_R^*$, yielding the bounded segment $s_d=\overline{\mathbf{Q}_L \mathbf{Q}_R}$ where $\mathbf{Q}_L$ and $\mathbf{Q}_R$ denote the intersection points of $\ell_d$ with $\ell_L^*$ and $\ell_R^*$ respectively. These calibrations were established in consultation with experts from the transit authority.

 Using the four platform corner points $\left(\mathbf{T}_L, \mathbf{T}_R, \mathbf{B}_L, \mathbf{B}_R\right)$, two internal boundaries are defined by selecting points at fixed fractions of the top and bottom edges ($\alpha=1 / 5$, $\beta=1 / 7$) and connecting them, resulting in the line segments, $\ell_{\text {left }}$ and $\ell_{\text {right }}$ (Fig. \ref{FIG:2}b). The resulting boundaries $\ell_{\text {left }}=\overline{\mathbf{T}_L^{(\alpha)} \mathbf{B}_L^{(\beta)}}$ and $\ell_{\text {right }}=\overline{\mathbf{T}_R^{(\alpha)} \mathbf{B}_R^{(\beta)}}$ will define the boundaries of wall-proximal and yellow-line-proximal zones.

The final three zones are defined as shown in Fig. \ref{FIG:2}c:
\begin{itemize}
    \item \textit{\textbf{Zone (A): Wall-Proximal Zone.}} Corresponds to the area near the platform wall, farthest from the track. It is defined by the polygon $\left(\mathbf{T}_L, \mathbf{T}_L^{(\alpha)}, \mathbf{B}_L^{(\beta)}, \mathbf{B}_L\right)$.


    \item \textit{\textbf{Zone (B): Yellow-Line Proximal Zone.}} Corresponds to a transitional space where passengers may approach the yellow line while remaining within designated safety boundaries. It is defined by the polygon $
    \left(\mathbf{T}_R^{(\alpha)},\mathbf{T}_R, \mathbf{B}_R,\mathbf{B}_R^{(\beta)}\right)$.


    \item \textit{\textbf{Zone (C): Platform Far-End Zone.}} Corresponds to the area immediately adjacent to the platform edge, it is the nearest area to the tunnel defined by the polygon $
\left(\mathbf{T}_L,\mathbf{T}_R, \mathbf{Q}_R,\mathbf{Q}_L\right)$.


\end{itemize}


\subsection{Position Risk Heatmap Generation}
\label{sec4.4}
We employ a homography transformation that maps image-plane coordinates to a platform-referenced coordinate system using the homography matrix $M$. This transformation allows tracked person positions to be expressed in a common spatial frame aligned with the actual platform geometry. As a result, spatial proximity, persistence, and movement patterns can be interpreted consistently across time and individuals.

\begin{figure}[t]
        \centering	\includegraphics[width=0.5\textwidth]{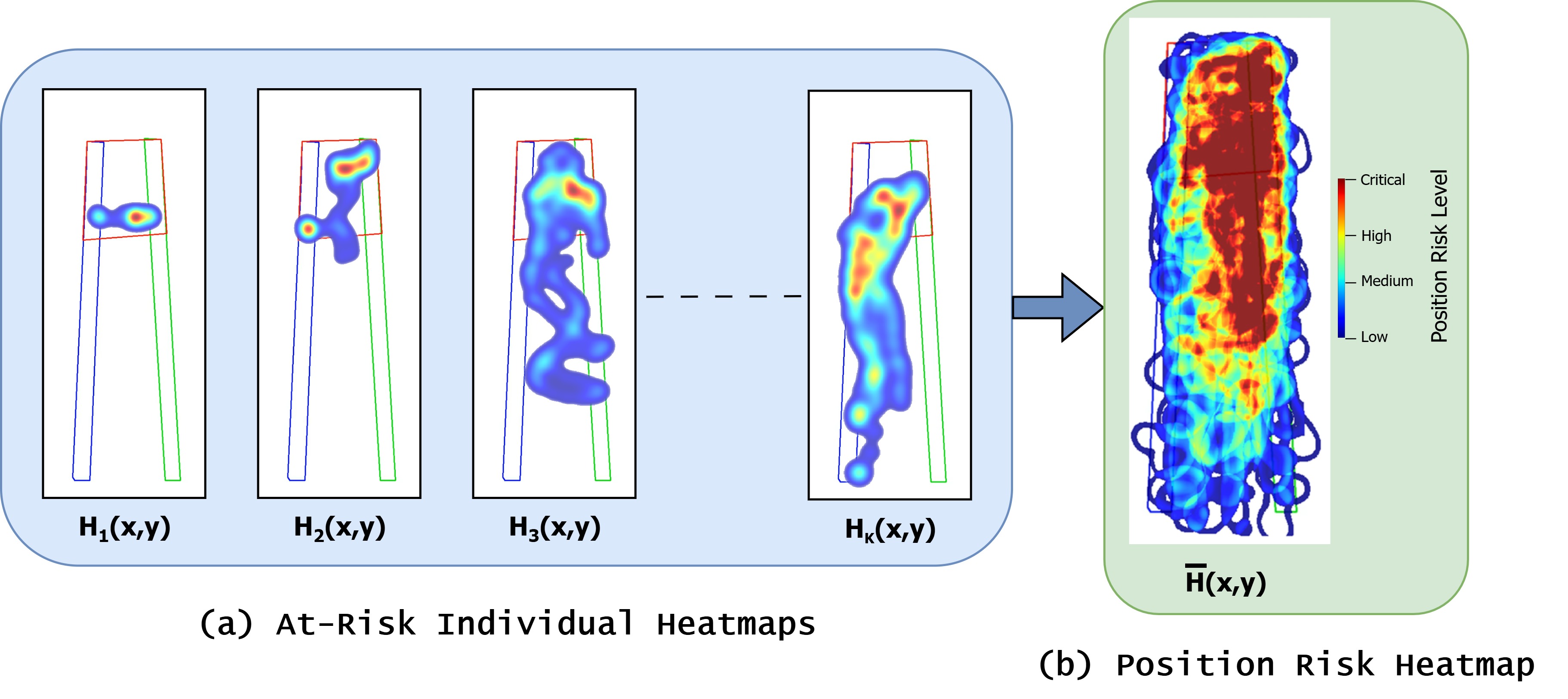}
	\caption{Illustration of Individual At-Risk Heatmaps and their aggregation into a Position Risk Heatmap.}
	\label{FIG:3}
\end{figure}

Once projected into the platform-referenced space, each tracked individual is represented by a sequence of two-dimensional positions over time. The heatmap of each tracked individual is generated mathematically through a density accumulation and Gaussian smoothing process over its 2D trajectory points as shown in Fig. \ref{FIG:3}a. We define a heat density map $\mathcal{H}(x, y)$ as:

\begin{equation}
\mathcal{H}(x, y)=\sum_{j=1}^N \delta\left(x-x_j, y-y_j\right)
\end{equation}
where $\left(x_j, y_j\right)$ are the recorded 2D trajectory points of a person,
$N$ denotes the number of recorded trajectory points for a given individual, and $\delta(\cdot)$ is the Dirac delta representing one unit of heat per position.

Individual heatmaps provide information about the spatial behavior of each person, but suicide risk in metro platforms is also shaped by persistent environmental factors and recurrent patterns. A global representation is therefore required to capture shared areas that emerge across different at-risk individuals.
To obtain this position risk heatmap, we aggregated $K$ at-risk individual trajectory-based heatmaps from our dataset. This aggregation yields a scene-level spatial representation reflecting the collective distribution of different at-risk persons on the platform. The position risk heatmap $
\overline{\mathcal{H}}
$ is computed as:
\begin{equation}
\overline{\mathcal{H}}(x, y)=\frac{1}{K} \sum_{k=1}^{K} \mathcal{H}_k(x, y)
\end{equation}

Regions that are repeatedly occupied or associated with prolonged dwell time naturally emerge in the heatmap as high-intensity areas, while transient regions remain low as depicted in Fig. \ref{FIG:3}b. 

The aggregated position risk heatmap $\overline{\mathcal{H}}(x,y)$ is normalized in the range $[0,1]$ to obtain a position risk score ${PR_p}$. This position risk score is later integrated to support the final suicide risk assessment stage.

\subsection{Suicide Risk Assessment}
\label{sec4.5}
Based on the behavioral studies for suicide prevention in metro stations discussed in Section \ref{sec2.1}, we derive eight interpretable suicide risk indicators for each person $p$ from the outputs of the previous sections:
\begin{enumerate}

\item \textbf{Position risk score ($PR_p$).} As described in Section \ref{sec4.4}, it is computed by averaging the position risk heatmap $\overline{\mathcal{H}}$ values along an individual’s trajectory, capturing whether a person occupies spatial locations associated with recurrent exposure and quantifying this occupation.
\item \textbf{Walk/Stand on the yellow line ($Cr_p$).} This binary indicator denotes whether an individual is detected performing a walking or standing action while positioned on the yellow line. It is activated when a person's left or right leg keypoint crosses the right boundary of Zone B and is simultaneously classified as walking or standing by the activity recognition module (Section \ref{sec4.2}).
\item \textbf{Yellow line crossings ($NCr_p$).} It counts the number of distinct events in which the right or left leg keypoints of an individual cross the
right boundary of Zone B. 
\item \textbf{Time on the yellow line ($TY_p$).} It accumulates the total duration spent on the yellow line region in seconds.
\item \textbf{Longest yellow line segment ($LY_p$).} It records the maximum continuous duration spent on the yellow line region without interruption.
\item \textbf{Back-and-forth between wall and yellow line count ($BF_p$).} It measures repeated transitions between Zone A and Zone B derived from Section \ref{sec4.3}. A transition is recorded when the right and left leg keypoints of an individual cross one of the two zones, move to the other zone, and subsequently return to the initial one: 'A → B → A' or 'B → A → B' transitions.
\item \textbf{Look Tunnel ($LT_p$).} This binary indicator is obtained from the activity recognition module (Section \ref{sec4.2}) and indicates whether the individual is detected repeatedly orienting his upper body toward the tunnel area. 
\item \textbf{Enter the far-end zone of the platform ($E_p$).} This binary indicator determines whether the individual has entered the platform far-end zone (Zone C).
\end{enumerate}
Given the set of risk indicators defined above, we assess suicide risk using an XGBoost supervised model \cite{Chen2016XGBoostAS} operating on accumulated features over a temporal window. We adopt XGBoost for its robustness under limited and imbalanced data conditions, making it well-suited for our individual-level suicide risk assessment \cite{velarde2023evaluatingxgboostbalancedimbalanced}. As defined in Eq. (\ref{eq:1}), for an individual $p$, a feature vector $\mathbf{x}_p(W_t) \in \mathbb{R}^d$ is extracted over $W_t$ as:

\begin{equation}
\begin{aligned}
\mathbf{x}_p(W_t)=\big[ &
PR_p(W_t),\; Cr_p(W_t),\; NCr_p(W_t),\; TY_p(W_t),\\
& LY_p(W_t),\; BF_p(W_t),\; LT_p(W_t),\; E_p(W_t)
\big]
\end{aligned}
\end{equation}

\noindent where each component corresponds to one of the proposed
suicide risk indicators accumulated within the window $W_t$. This feature vector is then provided as input to our XGBoost, which outputs a continuous suicide risk score $R_p(W_t)$ between 0 and 1.

\section{Experimental Evaluation}

\subsection{Dataset Description}

With approval from the Société de transport de Montréal (STM) and the Research ethic Institutional Review Board of our universities, we accessed surveillance video recordings of suicide attempts in the metro stations of Montréal. They provided 66 five-minute surveillance video recordings of cases before an attempt, and 56 recordings from the same cameras at the same time of day, and day of week, but when no suicide attempt occurred. 
For the suicide risk assessment task, we created a dataset with annotations provided at the individual level, including their extracted risk indicators, with the assistance of psychology experts and researchers in the field of suicidology. Individuals are labeled according to whether they originate from an at-risk or control scenario, without frame-level intent annotations. Using this procedure, we obtained a total of 256 individual instances, comprising 190 control subjects and 66 at-risk subjects. Data splits are performed at the video level, ensuring that all individuals extracted from the same recording belong to the same subset. The dataset was split into 75\% training and 25\% test, while preserving class proportions between at-risk and control.

\subsection{Implementation Details}

Our XGBoost model is trained with 300 boosting iterations and a learning rate of 0.05. Tree depth is limited to four to control model complexity, while subsampling of training instances and features is applied with ratios of 0.85 to improve generalization. Class imbalance between control and at-risk individuals is addressed by weighting the positive class proportionally to its inverse frequency in the training set. The model is optimized using the binary logistic objective and primarily evaluated using the ROC-AUC metric, which measures the ability of the model to discriminate between at-risk and control individuals independently of any decision threshold. In addition, we report Sensitivity, which reflects correct identification of at-risk cases, and Specificity which reflects correct identification of control cases. We also report the false positive rate (FPR), corresponding to false alarms, and the false negative rate (FNR), corresponding to missed detection rate. These metrics are computed by applying a fixed decision threshold of 
$R_p >$ 0.2, chosen to favor early detection of at-risk situations in line with suicide prevention objectives.

\subsection{Quantitative Results}

\begin{figure}[t]
        \centering	\includegraphics[width=0.5\textwidth]{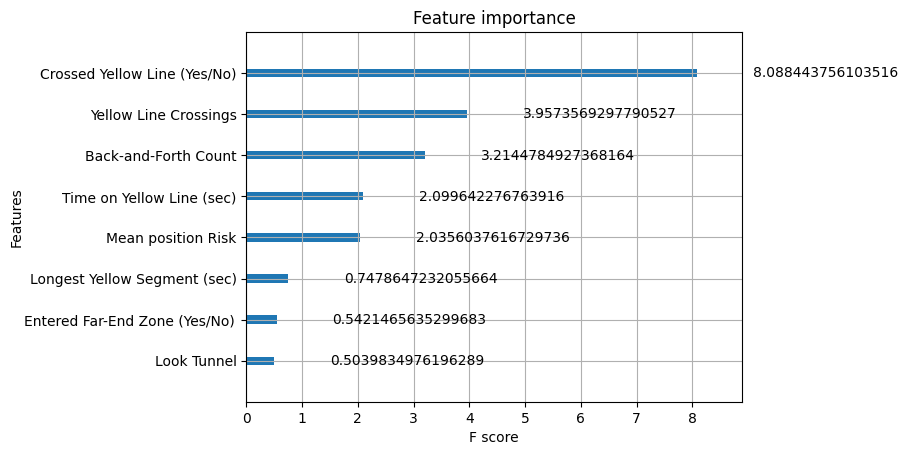}
	\caption{Feature importance of the eight risk indicators used for suicide risk assessment, computed from our trained XGBoost model. Higher F-scores indicate more frequent usage of a feature.}
	\label{FIG:4}
\end{figure}

Table \ref{tab:tab1} reports 10-Fold Cross-Validation results of the proposed suicide risk assessment framework under different controlled configurations, where detection/tracking and activity recognition modules are alternatively replaced by ground-truth annotations to assess the robustness of our framework and to quantify the impact of different modules on the final decision. The results show that replacing the predicted detection/tracking module in the fully automatic configuration with ground-truth annotations increases ROC–AUC from 0.832 ± 0.095 to 0.919 ± 0.054, while substantially improving sensitivity (0.631 → 0.800) and reducing false negative rates (0.369 → 0.200). In contrast, substituting the activity recognition module with ground-truth labels yields more moderate gains, with ROC–AUC increasing from 0.832 ± 0.095 to 0.893 ± 0.051. The remaining gap between fully automatic configuration (0.832 ± 0.095) and the upper bound configuration (0.924 ± 0.055) highlights room for future improvements to the indicator computation modules. As well, it highlights the need for researchers in suicidology to continue studying possible visual indicators that can be actionable with computer vision and machine learning algorithms.  

Beyond predictive performance, we analyze model interpretability to better understand the contribution of each risk indicator. Fig. \ref{FIG:4} reports the feature importance scores obtained from the trained XGBoost model for the fully automatic configuration, which reflects how frequently a feature is used to split the data across the ensemble of trees. The results show that indicators related to direct interaction with the yellow line play a dominant role in suicide risk assessment. In particular, crossing the yellow line emerges as the most influential feature ($\simeq$ 8.1 F-Score), followed by the number of yellow line crossings and the back-and-forth movement count. The time spent on the yellow line and the position risk score also contribute substantially, highlighting the importance of prolonged presence in high-risk regions rather than isolated events. Other behavioral cues, such as entering the far-end zone of the platform and looking at the tunnel, provide complementary information but are less decisive when considered independently. These findings are consistent with prior psychological and behavioral studies.

\begin{figure}[t]
        \centering	\includegraphics[width=0.5\textwidth]{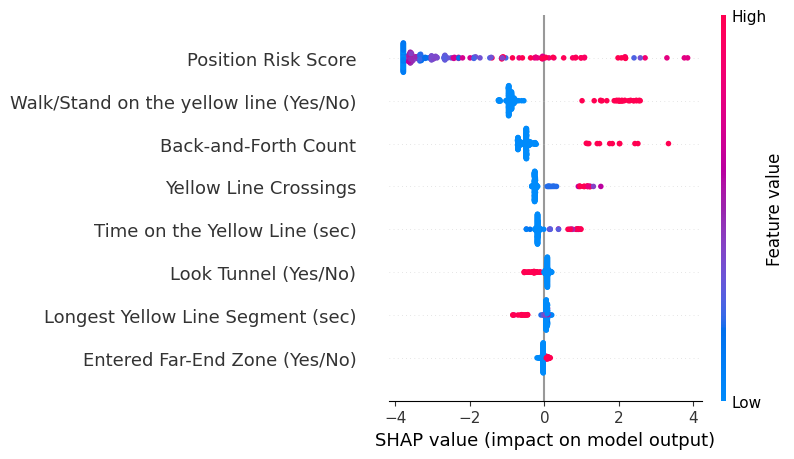}
	\caption{Model interpretability analysis using SHAP values, highlighting how each indicator influences suicide risk predictions.}
	\label{FIG:5}
\end{figure}

Fig. \ref{FIG:5} presents the SHAP summary plot, illustrating both the relative importance of each risk indicator and the directional influence of their values on the predicted suicide risk score. Consistent with the feature importance analysis, position risk score and direct interaction with the yellow line emerge as dominant contributors. Higher values of the position risk score are associated with positive SHAP values, indicating a strong increase in predicted risk when individuals occupy spatial regions frequently associated with prior at-risk trajectories. Similarly, crossing the yellow line shows a clear separation, where positive instances (high feature values) consistently push the suicide risk score upward. Indicators capturing repetitive and persistent behaviors, such as back-and-forth movements, number of yellow line crossings, and time spent on the yellow line, also exhibit predominantly positive SHAP contributions when their values increase. 

\begin{table*}[t]
  \caption{10-Fold Cross-Validation results of our SRA-Framework under different module configurations. In the GT-Assisted configurations, we replaced Detection/Tracking or Activity recognition modules with GT annotations. In the Upper Bound configuration, we exclusively used GT annotations for SRA. SD corresponds to the standard deviation.}
  \centering
\small
\begin{tabular}{|l|l |l|c|c|c|c|c|} \hline \multirow{2}{*}{\textbf{Configuration}} & \multicolumn{2}{c|}{\textbf{Modules}} & \multicolumn{ 5}{|c|}{ \textbf{Statistical Measures (± SD)} } \\ \cline{2-8} & \textbf{\textit{Detection / Tracking}} & \textbf{\textit{Act. Rec.}} & \textbf{\textit{ROC-AUC}} 
& \textbf{\textit{Sensitivity}}    
& \textbf{\textit{Specificity}}    
& \textbf{\makecell{\textit{False Alarm}\\\textit{Rate (FPR)}}}        
& \textbf{\makecell{\textit{Missed Detection}\\\textit{Rate (FNR)}}}

           \\
\hline
\hline
GT-Assisted &
YOLOx / Bytetrack &         GT & 0.893 ± 0.051 & 0.631 ± 0.228 & 0.913 ± 0.056 & 0.087 ± 0.056 & 0.369 ± 0.228 \\
GT-Assisted &
        GT &      SSTAR & 0.919 ± 0.054 & 0.800 ± 0.187 & 0.915 ± 0.046 & 0.085 ± 0.046 & 0.200 ± 0.187 \\
Upper Bound &
        GT &         GT & 0.924 ± 0.055 & 0.867 ± 0.113 & 0.929 ± 0.064 & 0.071 ± 0.064 & 0.133 ± 0.113 \\
Fully Automatic &
YOLOx / Bytetrack &      SSTAR & 0.832 ± 0.095 & 0.631 ± 0.228 & 0.902 ± 0.071 & 0.098 ± 0.071 & 0.369 ± 0.228 \\
\hline
\end{tabular}  
\label{tab:tab1}%
\end{table*}

\begin{figure*}
\begin{subfigure}{.5\textwidth}
  \centering
  \includegraphics[width=1.0\linewidth]{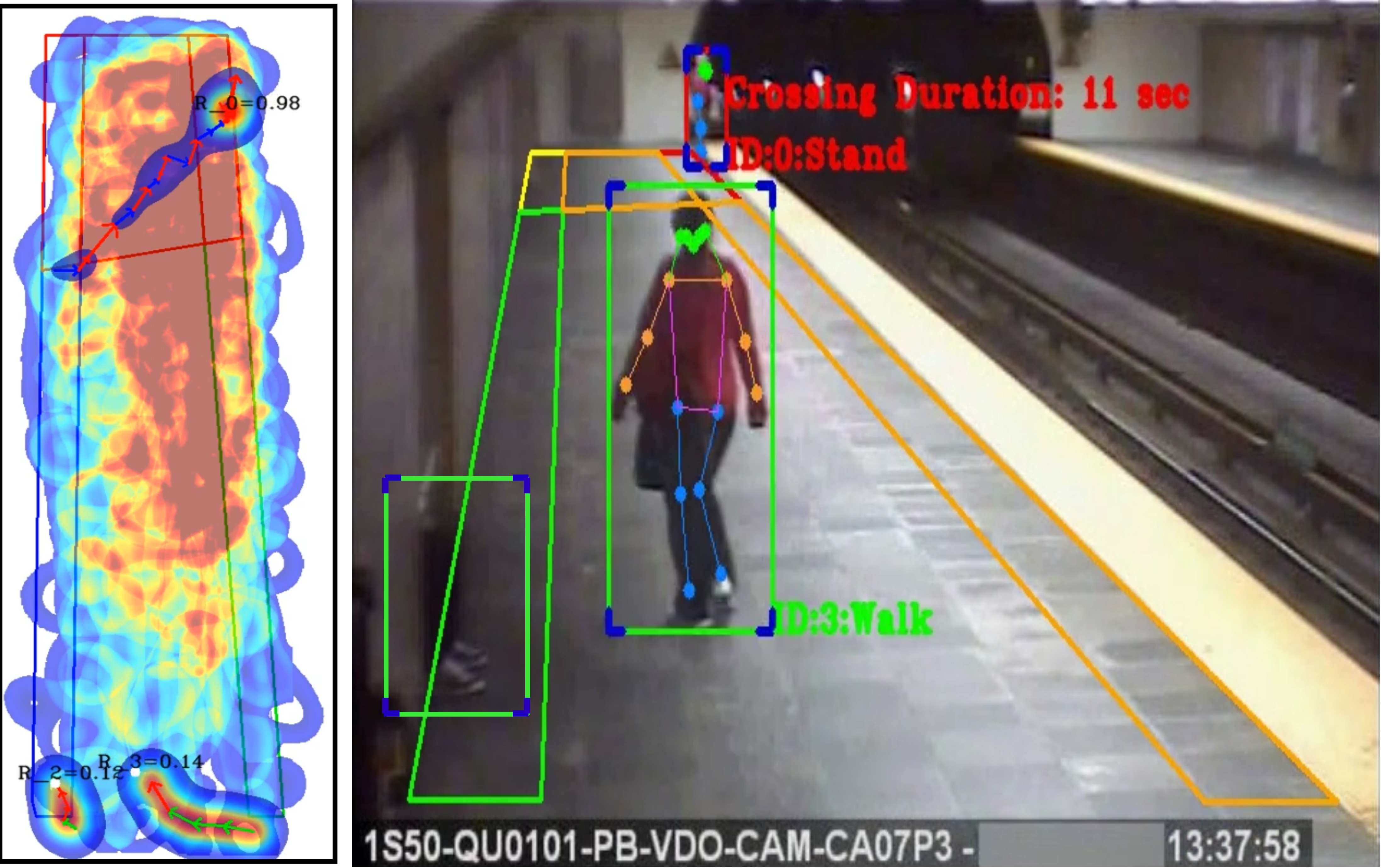}
  \caption{}
  \label{fig:sfig1}
\end{subfigure}%
\begin{subfigure}{.5\textwidth}
  \centering
  \includegraphics[width=1.0\linewidth]{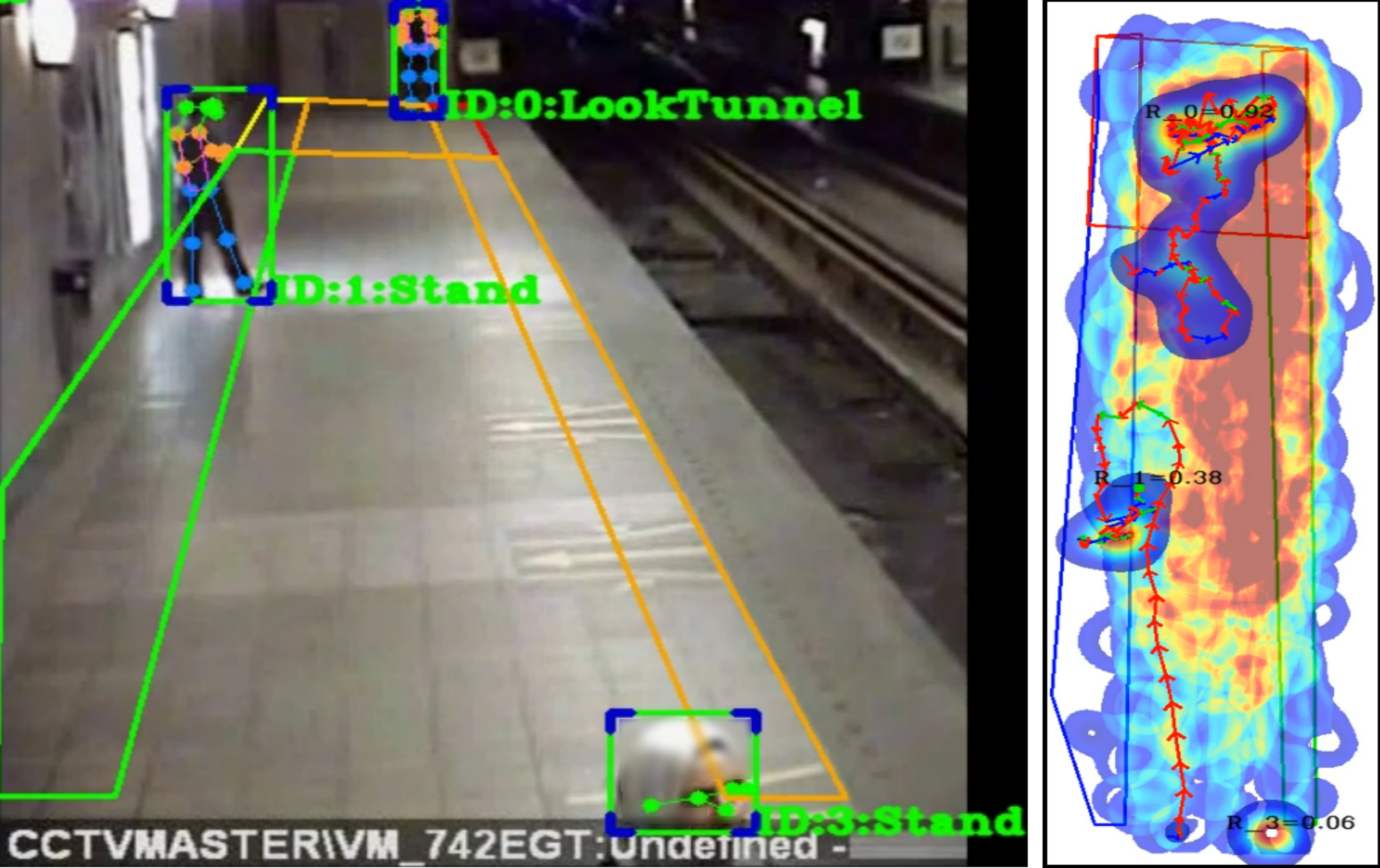}
  \caption{}
  \label{fig:sfig2}
\end{subfigure}
\caption{\textbf{Qualitative Results.} Output predictions of two frames from surveillance video streams.}
\label{FIG:6}
\end{figure*}

\subsection{Qualitative Results}

Fig. \ref{FIG:6} illustrates two platform states predicted by our SRA-Framework. The first video illustrated in Fig. \ref{fig:sfig1} presents three persons on the platform, our system correctly assigns a high risk score ($R_0$=0.98), driven by a combination of a prolonged crossing of the yellow line (11 seconds) and elevated position risk score along the position risk heatmap. Our system also successfully assigned lower suicide risk scores for control individuals ($R_2$=0.12 and $R_3$=0.14).
In the second video illustrated in Fig. \ref{fig:sfig2}, our system correctly assigns a high risk score of 
$R_0$=0.92 to the at-risk individual (ID 0), while correctly estimating substantially lower risk scores for control individuals ($R_1$=0.38 and $R_3$=0.06). This distinction arises from the accumulation of multiple risk indicators for ID 0, including prolonged crossing of the yellow line, a look-tunnel action, and sustained presence in spatial regions associated with elevated risk in the position risk heatmap. 

\section{Limitations and Future Work}
Despite encouraging results, several limitations should be acknowledged. First, the dataset used in this study is limited in size. Although this reflects real-world deployment conditions, larger and more diverse dataset would be beneficial. Second, for the proposed framework, errors introduced at early stages, particularly in detection, tracking, and activity recognition, can propagate to downstream risk indicators. Our controlled evaluations using ground-truth substitutions in Table \ref{tab:tab1} show that further improvements in module choice could significantly enhance the overall performance. Third, the set of risk indicators considered in our SRA-Framework does not cover all behavioral cues identified in psychological and behavioral studies, such as leaving an object on the platform. Waiting on the platform until one or more trains passed was not covered in our study because of using only 5-min videos before the train arrived.

Future research directions include improving tracking under crowded conditions and incorporating additional interpretable behavioral indicators. Ultimately, integrating human-in-the-loop feedback and evaluating the framework in real-time operational settings with the STM represent important steps toward practical deployment.

\section{Conclusion}

Our work introduces the first complete framework for Suicide Risk Assessment (SRA) in metro stations, integrating perception, behavioral analysis, spatial reasoning, and interpretable risk inference within a unified pipeline. Unlike prior works that address isolated components such as detection, tracking, or activity recognition, our approach benchmarks a full operational pipeline whose outputs directly support safety-oriented decision making in real-world surveillance environments. Beyond performance, our study emphasizes interpretability, showing that risk assessments are driven by intuitive indicators aligned with established behavioral and spatial risk factors. This positions the proposed framework as a meaningful bridge between AI-based surveillance systems and interdisciplinary research on suicide prevention.

\section*{Acknowledgements}

This work was supported by the Canada Research Chair in Artificial Intelligence for Suicide Prevention (CRC-2023-00036).

\section*{Ethical Statement}

This study was conducted in accordance with applicable ethical guidelines and received approval from the relevant institutional ethics committee. All video data were handled under confidentiality protocols and anonymized through systematic blurring of identifiable visual information prior to analysis. The data were used exclusively for suicide prevention research, securely stored, and accessible only to authorized researchers.
\newline
The literature documents differences in suicidal behaviors across sex and age groups, as well as algorithmic biases in computer vision systems related to individuals’ characteristics. However, the size of our dataset and the quality of the recordings did not allow for a differentiated analysis by sex and age to assess potential biases. In future work, access to a larger dataset collected with a new generation of cameras will enable disaggregated analyses by sex and age groups to examine behavioral variations.

\bibliographystyle{named}
\bibliography{ijcai26}

\end{document}